\pdfoutput=1

\documentclass[11pt]{article}

\usepackage[final]{acl}

\usepackage{times}
\usepackage{latexsym}

\usepackage[T1]{fontenc}

\usepackage[utf8]{inputenc}

\usepackage{microtype}

\usepackage{inconsolata}

\usepackage{graphicx}
\usepackage{hyperref}
\usepackage{url}
\usepackage{booktabs} 
\usepackage{tabularx}
\usepackage{multicol}
\usepackage{multirow}
\usepackage{subcaption}
\usepackage{colortbl}
\usepackage{array}
\usepackage{algorithm}
\usepackage{algpseudocode}
\usepackage{amsmath}
\usepackage{amsfonts}
\usepackage{amssymb}

\title{WaterSeeker: Pioneering Efficient Detection of Watermarked Segments in Large Documents}

\author{
Leyi Pan\textsuperscript{1}\thanks{Work was done during the intern at Zhipu AI.},
~~~ Aiwei Liu\textsuperscript{1},
~~~ Yijian Lu\textsuperscript{1},
~~~ Zitian Gao\textsuperscript{2},
~~~ Yichen Di\textsuperscript{1}, \\ 
~~~ \bf{Shiyu Huang}\textsuperscript{3}, 
~~~ Lijie Wen\textsuperscript{1}\thanks{Corresponding author}, 
~~~ Irwin King\textsuperscript{4},
~~~ Philip S. Yu\textsuperscript{5}\\
\textsuperscript{1}Tsinghua University~~~
\textsuperscript{2}The University of Sydney~~~
\textsuperscript{3}Zhipu AI~~~\\
\textsuperscript{4}The Chinese University of Hong Kong~~~
\textsuperscript{5}University of Illinois at Chicago~~~\\
{\tt\small panly24@mails.tsinghua.edu.cn, liuaw20@mails.tsinghua.edu.cn, wenlj@tsinghua.edu.cn}
}

\begin{document}
\maketitle
\begin{abstract}
Watermarking algorithms for large language models (LLMs) have attained high accuracy in detecting LLM-generated text. However, existing methods primarily focus on distinguishing fully watermarked text from non-watermarked text, overlooking real-world scenarios where LLMs generate only small sections within large documents. In this scenario, balancing time complexity and detection performance poses significant challenges. This paper presents WaterSeeker, a novel approach to efficiently detect and locate watermarked segments amid extensive natural text. It first applies an efficient anomaly extraction method to preliminarily locate suspicious watermarked regions. Following this, it conducts a local traversal and performs full-text detection for more precise verification. Theoretical analysis and experimental results demonstrate that WaterSeeker achieves a superior balance between detection accuracy and computational efficiency. Moreover, its localization capability lays the foundation for building interpretable AI detection systems. Our code is available at \href{https://github.com/THU-BPM/WaterSeeker}{https://github.com/THU-BPM/WaterSeeker}.
\end{abstract}

\section{Introduction}
\label{sec:intro}
\begin{figure*}
    \centering
    \includegraphics[width=\linewidth]{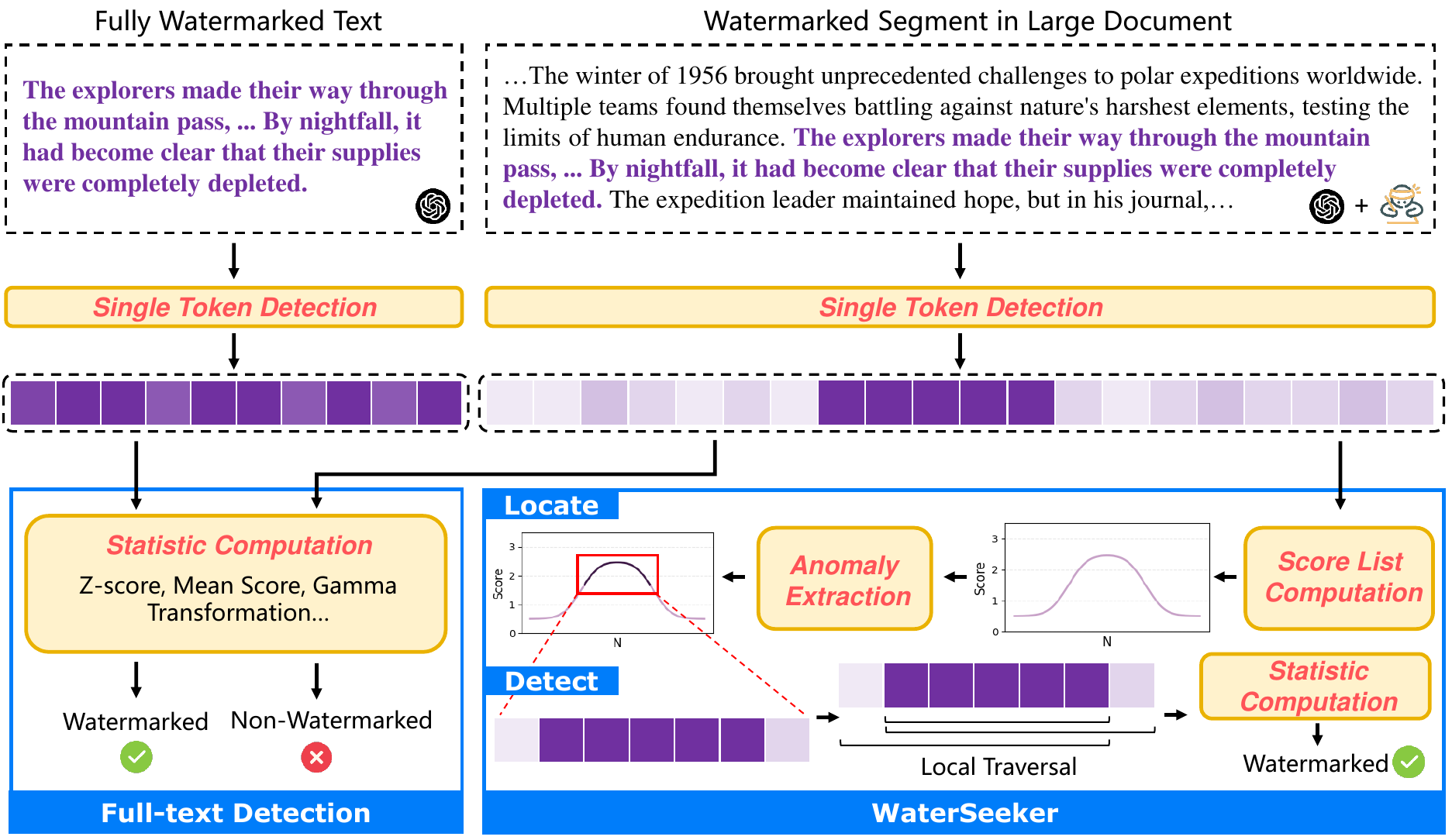}
    \caption{While full-text detection methods effectively differentiate between fully watermarked and non-watermarked texts, they often struggle with watermarked segment detection due to the dilution effect. To address this, WaterSeeker employs a "first locate, then detect" strategy, which narrows the detection range before conducting local traversal for further verification.}
    \label{fig:overview}
\end{figure*}

As large language models (LLMs) generate high-quality text, they address practical challenges but also raise concerns such as misinformation \citep{megias2022architecture,chen2023can} and copyright infringement \citep{rillig2023risks}. LLM watermarking technology has emerged to tackle these issues by embedding specific information (watermarks) during text generation, allowing for accurate detection through specialized algorithms. Current watermark detection methods \citep{DBLP:conf/icml/KirchenbauerGWK23,zhao2023provable,liu2023semantic,aronsonpowerpoint,lu2024entropy,lee2023wrote,hu2023unbiased,wu2023dipmark} first calculate watermark scores for individual tokens through single token detection, then compute statistics across the entire document for classification. While these full-text detection methods effectively distinguish between fully watermarked and non-watermarked texts, they fail in real-world scenarios where LLMs generate only brief segments within longer documents, due to dilution effects, as shown in Figure \ref{fig:overview}. To the best of our knowledge, the WinMax algorithm \citep{kirchenbauer2023reliability} is the only work addressing this limitation by examining all possible window sizes and selecting the maximum statistical score across all windows as the final detection result, but suffers from high time complexity.

To address these issues, we propose a novel and general watermark detection method called WaterSeeker. WaterSeeker follows a "first locate, then detect" approach, as shown in Figure \ref{fig:overview}. It initially employs a low-complexity anomaly points extraction algorithm to identify suspected watermark regions, narrowing the detection target from a long text to a small segment encompassing the ground truth segment. Next, a local traversal is performed on the localization result, conducting full-text watermark detection within each window and comparing the highest confidence result to a threshold for the final determination. Theoretical analysis suggests that this coarse-to-fine process has the potential to achieve optimal detection performance while maintaining the lowest possible complexity for solving this problem.

In the experiment, we compared the effectiveness and time complexity of WaterSeeker with baseline methods for detecting watermarked segments in large documents. WaterSeeker significantly outperformed the baselines in balancing time complexity and detection performance. Moreover, it demonstrates good adaptability to different watermark strengths, segment lengths, and document lengths, while achieving robust performance against text edit attacks. In summary, the contributions of this work are as follows: 
\begin{itemize}
    \item We comprehensively define a new scenario: detecting watermarked segments in large documents. This includes specifying algorithm inputs/outputs, evaluation metrics, and how to create test datasets.
    \item We propose WaterSeeker, a general watermark detection method that effectively identifies watermarked segments in large documents, tackling the issues caused by dilution effects.
    \item WaterSeeker outperforms baselines in achieving a superior balance between time complexity and detection effectiveness.
    \item Further experiments demonstrate that WaterSeeker exhibits strong adaptability across various watermark strengths, segment lengths, and document lengths, while also being robust against text editing attacks.
\end{itemize}
\section{Related Work}
\label{sec:related}
Currently, mainstream LLM watermarking methods involve modifying the inference phase by altering logits or influencing token sampling \citep{liu2023survey,pan2024markllm, liu2024prevent}. The KGW family \citep{DBLP:conf/icml/KirchenbauerGWK23,zhao2023provable,hu2023unbiased,liu2023semantic,wu2023dipmark,liu2023private,he2024can,huo2024tokenspecific} categorizes vocabulary into green and red lists, biasing towards green tokens during generation. The bias value is typically determined by the parameter $\delta$, which reflects the watermark strength. For these methods, single token detection determines whether each token belongs to the green list, while full-text detection involves calculating the z-score of green tokens across the entire document; exceeding a threshold indicates watermarking.

On the other hand, the Aar family \citep{aronsonpowerpoint,christ2024undetectable,kuditipudi2023robust} uses pseudo-random sequences to guide token sampling. It generates a pseudo-random vector \( u \sim \text{Uniform}([0, 1])^{|V|} \) based on previous tokens and selects the token $i$ maximizing \( u_i^{1/p_i} \), where $p$ is the LLM's probability vector. Watermark strength is controlled by sampling temperature. In these methods, single token detection calculates the correlation value between each token and \(u\), while full-text detection applies gamma transformation to derive the detection confidence. Details of the KGW and Aar watermarking algorithms can be found in Appendix \ref{sec:appendix_scheme}.

Despite the high accuracy of watermarking algorithms for distinguishing between fully watermarked and non-watermarked text, their performance fall sharply when detecting watermarked segments within large documents. A few studies have mentioned copy-paste attack \citep{kirchenbauer2023reliability,yoo2023advancing,wang2023towards}, which involves mixing a portion of watermarked text with non-watermarked content, similar to our scenario. \citet{yoo2023advancing} and \citet{wang2023towards} evaluated their methods' robustness against copy-paste attacks by combining 10\% to 50\% watermarked text with non-watermarked text. However, as they did not develop specific detection mechanisms for this situation, their findings showed that their methods were not robust against this type of attack.

Among existing studies, only WinMax \citep{kirchenbauer2023reliability} specifically addresses watermarked segment detection by examining all possible window sizes to find the highest local statistics. However, its high time complexity limits practical application. To address this limitation, we propose WaterSeeker, a novel method that employs a "first locate then detect" strategy to achieve efficient detection of watermarked segments in large documents.
\section{Problem Formulation}
\label{sec:problem}
\textbf{Definition}. Given a text of length $N$ containing $m$ watermarked segments at position $[(s_1, e_1), ... (s_m, e_m)]$, the objective is to determine the presence and location of the watermarked segment. The detection algorithm outputs: \{`has\_watermark': True/False, `indices': $[(s_1', e_1'), ..., (s_{m'}', {e}_{m'}')]$\}.

\vspace{3pt}

\noindent\textbf{Evaluation}. A watermark is considered successfully detected if: \textbf{(1)} \( \text{output}.\text{has\_watermark} = \text{True} \). \textbf{(2)} The overall Intersection over Union (IoU) between the detected segments ${(s'_i, e'_i)}|_{i=1}^{m'}$ and the ground truth segments $(s, e)|_{i=1}^m$ is positive, indicating no complete false detection: 

    {\small
    \begin{equation}
        \text{IoU} = \cfrac{L_{\text{intersection}}}{L_{\text{union}}} > 0.
    \end{equation}
    }
\noindent Based on these criteria, we evaluate the classification performance using False Positive Rate (FPR), False Negative Rate (FNR) and F1 Score, as well as the localization performance using average IoU between detected and ground truth segments.
\section{Baseline Methods}
\label{sec:baselines}
\noindent\textbf{Full-text Detection}. As explained in Section \ref{sec:intro}, involves calculate watermark scores for individual tokens and compute statistics across the entire documents for classification. 

\vspace{3pt}

\noindent\textbf{WinMax} \citep{kirchenbauer2023reliability} involves iterating through all possible window sizes, and for each window size, the entire text is traversed to compute statistics for each local window, taking the maximum score for final results. The detection process can be described by the following formula:

{\small
\begin{equation}
    \text{score} = \max_{1 \le w \le N} \max_{1 \le i \le N-w+1} F(x_{i:i+w-1}),
    \label{eq:winmax}
\end{equation}
}

where \( w \) is the length of the local window, \( x \) represents the text tokens, and \(F\) is the statistical function. The time complexity is evidently $O(N^2)$. We also introduce a WinMax variant where window size increases by intervals $> 1$, reducing complexity to $O(N^2/\text{interval})$.

\vspace{3pt}

\noindent\textbf{Fix-Length Sliding Window (FLSW)} is a self-constructed method that uses a fixed-length window to traverse the text. The text is flagged as watermarked if any statistic score within the local windows exceeds the threshold. The pseudocode for all baselines can be found at Appendix \ref{sec:appendix_baselines}.

\section{Proposed Method: WaterSeeker}
\label{sec:seeker}

\subsection{Theoretical Basis: Gold Index is the Best}
\label{sec:theory}

\begin{figure}[t]
    \centering
    \includegraphics[width=\linewidth]{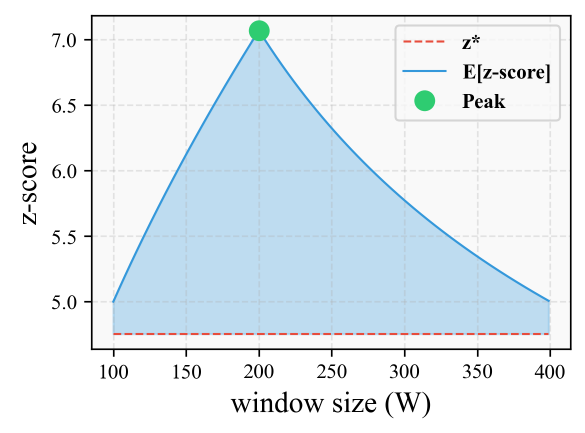}
    \caption{Expected z-score and the corresponding threshold $z^*$ across various $W$, $\alpha=10^{-6}, \gamma=0.5, \gamma_1=0.75$.}
    \label{fig:kgw_gold_best}
\end{figure}

This section provides the theoretical foundation of WaterSeeker, showing that using actual start and end indices (gold index) for watermark detection achieves the highest expected detection rate. We analyze using KGW \citep{DBLP:conf/icml/KirchenbauerGWK23} as a case study. Assuming $\gamma_1 > \gamma$ is the proportion of green tokens in the watermarked part, the watermark segment has a length $L$, and the statistical function $F$ is the z-score computation (detailed in Appendix \ref{sec:appendix_scheme}). Let's analyze the effect of window size $W$ on this statistic: (1) When $W < L$:

    \vspace{-10pt}
    
    {\small
    \begin{equation}
        E[z_W] = \frac{W\gamma_1 - \gamma W}{\sqrt{\gamma(1-\gamma)W}} = \sqrt{W} \cdot \frac{\gamma_1 - \gamma}{\sqrt{\gamma(1-\gamma)}}. 
    \end{equation}
    }

    \vspace{-5pt}
    
    (2) When $W > L$:

    \vspace{-7pt}
    
    {\small
    \begin{equation}
        E[z_W] = \frac{L\gamma_1 + (W-L)\gamma - \gamma W}{\sqrt{\gamma(1-\gamma)W}} = \frac{L(\gamma_1 - \gamma)}{\sqrt{\gamma(1-\gamma)W}}. 
    \end{equation}
    }
\vspace{-7pt}

From this, we can conclude that \textbf{when $W=L$, the z-score reaches its maximum}. During detection, we aim to maximize positive case z-scores while maintaining false positive rate below a target threshold $\alpha$. For a window of size $W$, the number of green tokens follows $B(W, \gamma)$, which approximates to $N(W\gamma, W\gamma(1-\gamma))$ for large $W$. This yields $z \sim N(0, 1)$, and the threshold $z^*$ is derived as: $z^* = \Phi^{(-1)}(1 - \alpha)$, which is a constant value for different $W$. The simulation result using real data are shown in Figure \ref{fig:kgw_gold_best}.

\subsection{Suspicious Region Localization}
\label{sec:locate}
Based on the theoretical analysis, WaterSeeker employs a coarse-to-fine process to gradually approximate the gold index. In the coarse step, a localization algorithm identifies suspicious regions and narrows detection to a small segment containing the ground truth, while maintaining minimal deviation. This step involves three sub-steps:

\vspace{3pt}

\noindent\textbf{(1) Score List Computation}: Similar to existing methods, watermark scores for individual tokens are calculated through single token detection. Then, a small sliding window (i.e. $W=50$) is used to traverse the text to compute average scores within the window, serving as a smoothing operation. This results in a score list $s$ of length $N-W+1$, where $s_i$ represents the average watermark intensity from text token $x_i$ to $x_i + W$.
    
\noindent\textbf{(2) Anomaly Extraction}: We design an anomaly extraction algorithm inspired by previous work in the field of style change detection \citep{zangerle2021overview} and intrinsic plagiarism detection \citep{manzoor2023exploring}. The mean score ($s_{\text{mean}}$) and top-k mean ($s_{\text{top-k-mean}}$) are calculated. Outliers are determined by:

    {\small
    \begin{equation}
    \text{score} > s_{\text{mean}} + \frac{(s_{\text{top-k-mean}} - s_{\text{mean}})}{2}.
    \end{equation}
    }
    
This is non-trivial because it ensures that the extracted suspicious watermarked regions likely cover the actual segments, with starting and ending deviations within a window size: when the sliding window falls entirely within the watermarked segment, the scores stabilize near $s_{\text{top-k-mean}}$; when the window falls completely outside, the scores stabilize below $s_{\text{mean}}$. Consequently, the extracted abnormal segment's start and end points ($s'$ and $e'$) generally satisfy $s' \in (s - W, s)$ and $e' \in (e, e + W)$. Moreover, the use of $s_{\text{top-k-mean}}$ ensures good adaptability to various watermark strengths. The corresponding experiment results can be found at Section \ref{sec:ablation}.

\noindent\textbf{(3) Fragment Connection}: Adjacent outliers are connected with a predefined tolerance threshold, and segments shorter than a minimum length are filtered out, producing a list of segment indices.

\vspace{-3pt}

\subsection{Local Traverse Detection}
\label{sec:detect}
After obtaining the coarse localization results, a fine-grained detection is performed by traversing segments within the predicted ranges. For each $(s', e')$ pair in the localization results, the algorithm examines segments with start points in $[s', s' + W)$ and end points in $(e' - W, e']$. Based on the previous analysis, these ranges likely contain the true watermarked indices. Full-text detection is performed on these segments, and the most significant statistic is compared against a threshold. The complete WaterSeeker algorithm is presented in Algorithm \ref{alg:seeker}.

\begin{algorithm}[t]
\caption{WaterSeeker Algorithm}
{
\begin{algorithmic}[1]
\Procedure{Localization}{tokens}
\State scores = SlidingWindow(tokens)
\State threshold = $s_{\text{mean}} + \frac{(s_{\text{top-k-mean}} - s_{\text{mean}})}{2}$
\State\Return ConnectOutliers(scores>threshold)
\EndProcedure

\Procedure{Detection}{tokens, segs}
\State detected = []
\For{$(s', e')$ in segs}
\State best = $-\infty$, idx = None
\For{$s \in [s', s'+W)$, $e \in (e'-W, e']$}
\State score=WatermarkScore(tokens[$s$:$e$])
\If{score > best}
\State best = score
\State idx = $(s,e)$
\EndIf
\EndFor
\If{best > threshold} 
    \State detected.append(idx)
\EndIf
\EndFor
\State\Return bool(detected), detected
\EndProcedure

\Procedure{WaterSeeker}{tokens}
\State\Return Detection(tokens, Localization(tokens))
\EndProcedure
\end{algorithmic}
}
\label{alg:seeker}
\end{algorithm}

\subsection{Time Complexity Analysis}
\label{sec:time_analysis}
\textbf{Time Complexity of WaterSeeker}. WaterSeeker consists of two main components: (1) Suspicious Region Localization with $O(N)$ complexity, where $N$ is the text length, and (2) Local Traverse Detection with $O(W^2)$ complexity, where $W$ is the window size. The total complexity is $O(N + W^2)$. In practice, $W^2$ is typically kept lower than $N$, as a slightly larger window (i.e., $W=50$, detailed in Appendix \ref{sec:appendix_window_size}) suffices for a smooth and low-noise representation of the surrounding watermark intensity. Thus, the overall time complexity of WaterSeeker is $O(N)$. 

\vspace{3pt}

\noindent\textbf{Lower Bound Complexity for the Problem}. To detect watermarked segments in a long text, any algorithm must examine each token in the text at least once. This requirement establishes a lower bound of $\Omega(N)$ for the time complexity of the problem, as at least one full pass through the text is necessary. Consequently, the WaterSeeker algorithm achieves a time complexity that matches the theoretical lower bound of the problem.

\section{Experiment}

\subsection{Experiment Settings}
\noindent\textbf{Watermarking Methods and Language Models}: We selected two representative watermarking algorithms, KGW \citep{DBLP:conf/icml/KirchenbauerGWK23} and Aar \citep{aronsonpowerpoint}, each at three strength levels. KGW's strength was set by the $\delta$ parameter (2.0=strong, 1.5=medium, 1.0=weak), while Aar's strength used the temperature parameter (0.5=strong, 0.4=medium, 0.3=weak). We used Llama-2-7b \citep{touvron2023llama} and Mistral-7b \citep{jiang2024mixtral} as generation models.

\vspace{3pt}

\noindent\textbf{Dataset Construction}: The first 30 tokens of each entry in the C4 dataset \citep{raffel2020exploring} were used for prompts. Watermarked segments of random length (100 to 400 tokens) were then generated using randomly selected watermark strengths. For positive examples, one such segment was randomly inserted into each 10,000-token Wikipedia passage \citep{wikidump}. While most experiments used single-segment insertion, we also conducted experiments with multiple watermarked segments inserted into the same passage, with results reported in Table \ref{tab:more_seg}. Negative examples consist of unmodified 10,000-token Wikipedia corpus. Based on this procedure, four datasets were created, each containing 300 positive and 300 negative examples: KGW-Llama, KGW-Mistral, Aar-Llama, and Aar-Mistral. 

\vspace{3pt}

\noindent\textbf{Baselines}: As introduced in Section \ref{sec:baselines}, we selected Full-text Detection and WinMax \citep{kirchenbauer2023reliability} with varying window size intervals, along with the Fixed-Length Sliding Window method using $W$ of 100, 200, 300, and 400.

\vspace{3pt}

\noindent\textbf{Hyper-parameters}: The parameters related to WaterSeeker are as follows: $W = 50, k=20$, with a tolerance for fragment connection set to 100. The threshold selection within the specified window is detailed in Appendix \ref{sec:appendix_threshold}. Notably, careful threshold selection is crucial for maintaining an acceptable false positive rate, as traversing long texts is prone to accumulating false positives. 

\subsection{Results of Full-text Detection}
\begin{table}[t]
    \caption{Results of full-text detection methods.}
    \centering
    \begin{tabular}{lccc}
        \toprule
        \textbf{Dataset} & \textbf{FPR$\downarrow$} & \textbf{FNR$\downarrow$} & \textbf{F1$\uparrow$} \\
        \midrule
        KGW-Llama & 0.000 & 0.983 & 0.033 \\
        KGW-Mistral & 0.000 & 0.993 & 0.013\\
        Aar-Llama & 0.000 & 0.980 & 0.039 \\
        Aar-Mistral & 0.000 & 0.980 & 0.039 \\
        \bottomrule
    \end{tabular}
    \label{tab:full_text}
\end{table}
Table \ref{tab:full_text} shows that full-text detection methods perform poorly across all four datasets, with an F1 score of less than 0.1. This indicates that full-text detection methods are totally ineffective for detecting watermarked segments in large documents.

\subsection{WaterSeeker Compared with WinMax}
\begin{table*}[t]
\caption{We evaluated the detection performance of WaterSeeker against various methods, including Full-text Detection, WinMax \citep{kirchenbauer2023reliability}, and FLSW. Results in this table utilized the Llama-2-7b model; results for Mistral-7b are presented in Appendix \ref{sec:appendix_main_mistral}. The metrics reported include false positive rate (FPR), false negative rate (FNR), F1 score, average Intersection over Union (IoU) between detected and ground truth segments, and processing time per sample. Best performances are highlighted in bold, while the second-best are underlined.}
\centering
\resizebox{0.98\textwidth}{!}{
\begin{tabular}{lccccc|ccccc}
\toprule
\multirow{2}{*}{\textbf{Method}} & \multicolumn{5}{c}{\textbf{KGW}} & \multicolumn{5}{c}{\textbf{Aar}} \\
\cmidrule{2-11}
& FPR$\downarrow$ & FNR$\downarrow$ & F1$\uparrow$ & IoU$\uparrow$ & Time(s)$\downarrow$ & FPR$\downarrow$ & FNR$\downarrow$ & F1$\uparrow$ & IoU$\uparrow$ & Time(s)$\downarrow$ \\
\midrule
WinMax-1 & 0.017 & \textbf{0.193} & \textbf{0.885} & \textbf{0.713} & 1632.11 & 0.017 & \textbf{0.277} & \textbf{0.831} & \textbf{0.616} & 3615.42\\
WinMax-50  & 0.017& 0.220& 0.868 & 0.672 & 34.31 & 0.007  & 0.307 & 0.816 & 0.577 & 72.13\\
WinMax-100 & 0.013 & 0.237 & 0.859 & 0.632 & 17.16 & \underline{0.003} & 0.330 & 0.800 & 0.554 & 35.34\\
WinMax-200  & 0.010 & 0.273 & 0.834 & 0.547 & 9.12 & \underline{0.003} & 0.363 & 0.776 & 0.486 & 18.38\\
\midrule
\cellcolor{gray!25}WaterSeeker  & \cellcolor{gray!25}0.017& \cellcolor{gray!25}\underline{0.213}& \cellcolor{gray!25}\underline{0.872}& \cellcolor{gray!25}\underline{0.675}& \cellcolor{gray!25}\textbf{1.75} & \cellcolor{gray!25}0.010& \cellcolor{gray!25}\underline{0.300}& \cellcolor{gray!25}\underline{0.819}& \cellcolor{gray!25}\underline{0.578}& \cellcolor{gray!25}\textbf{0.41}\\
\midrule
FLSW-100 & \textbf{0.003} & 0.383 & 0.761 & 0.451 & \underline{1.76} & \underline{0.003} & 0.440 & 0.716 & 0.403 & 1.31\\
FLSW-200  & \textbf{0.003} & 0.300 & 0.822 & 0.487 & \underline{1.76} & \textbf{0.000} & 0.380 & 0.765 & 0.411 & 1.29 \\
FLSW-300 & \underline{0.007} & 0.340 & 0.792 & 0.383 & \underline{1.76} & \textbf{0.000} & 0.413& 0.739 & 0.306 & 1.29 \\
FLSW-400  & \textbf{0.003} & 0.407 & 0.743 & 0.275 & \textbf{1.75} & \textbf{0.000} & 0.443 & 0.715& 0.228&   \underline{1.27}\\
\bottomrule
\end{tabular}
}
\label{tab:main}
\end{table*}

\vspace{3pt}

\begin{figure*}[t]
    \centering
    \includegraphics[width=\linewidth]{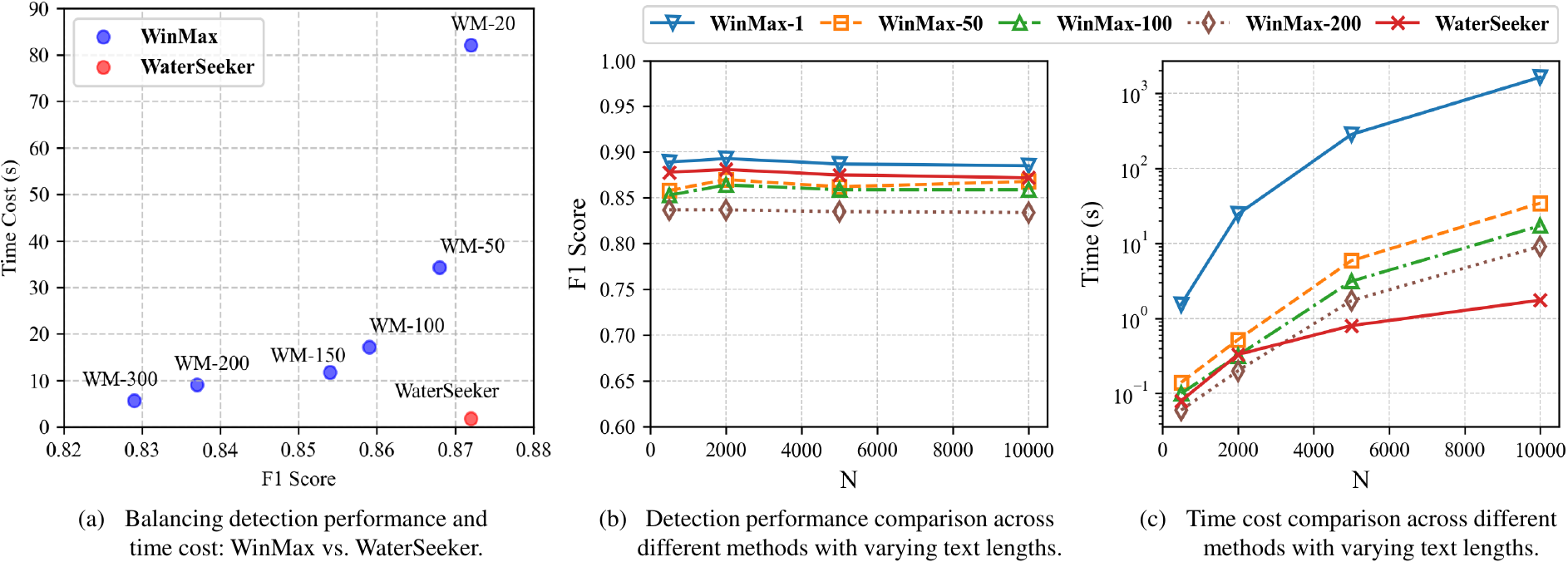}
    \caption{A detailed comparison of WinMax and WaterSeeker regarding their detection performance and time cost, as well as their performance across varying text lengths.}
    \label{fig:three-pic}
\end{figure*}
\noindent\textbf{Main Experiment}. From the data in Table \ref{tab:main}, we can compare the detection and localization capabilities of WinMax with different window size intervals (1, 50, 100, 200) and WaterSeeker for watermarked segments, as well as the time costs for processing individual samples. WaterSeeker's detection and localization performance is only slightly behind that of WinMax-1, while achieving a time savings of 1000 times. Given that WinMax evaluates all possible windows to ensure it reaches the gold index, it represents the upper bound of detection performance. However, as the window interval sizes for WinMax increase, the processing time decreases linearly with the interval size, yet it remains higher than that of WaterSeeker, while its detection and localization performance falls below that of WaterSeeker. \textit{Note: Table \ref{tab:main} demonstrates the results using Llama-2-7b, results for Mistral-7b can be found in Appendix \ref{sec:appendix_main_mistral}, which shows consistent trend.}

\vspace{3pt}

\noindent\textbf{Balancing Performance and Time Cost}. To clearly compare the balance between detection performance and time cost for WinMax and WaterSeeker, we collected additional data points for WinMax, as shown in Figure \ref{fig:three-pic}(a). Points further to the bottom right in the figure indicate a superior balance. It is evident that WaterSeeker is positioned clearly below and to the right of the curve formed by the WinMax data points, demonstrating a better balance between detection performance and time cost.

\vspace{3pt}

\noindent\textbf{Robustness against Varying Text Lengths}. To further validate the robustness of watermark detection algorithms against varying document lengths (mixing ratios), we tested WaterSeeker and WinMax at $N =$ 500, 2000, 5000, and 10000, measuring the detection F1 score and time cost, as illustrated in Figure \ref{fig:three-pic}(b), (c). Both WaterSeeker and WinMax exhibit stable detection performance with changes in $N$. However, WinMax’s time cost increases at a higher rate than that of WaterSeeker, indicating its impracticality for large documents.

\begin{table*}[t]
\caption{Comparison of detection and localization performance between WaterSeeker and FLSW with multiple segment insertion (one document containing three watermarked segments), including the time cost for processing each sample.}
\centering
\resizebox{0.98\textwidth}{!}{
\begin{tabular}{lccccc|ccccc}
\toprule
\multirow{2}{*}{\textbf{Method}} & \multicolumn{5}{c}{\textbf{KGW}} & \multicolumn{5}{c}{\textbf{Aar}} \\
\cmidrule{2-11}
& FPR$\downarrow$ & FNR$\downarrow$ & F1$\uparrow$ & IoU$\uparrow$ & Time(s) $\downarrow$ & FPR$\downarrow$ & FNR$\downarrow$ & F1$\uparrow$ & IoU$\uparrow$ & Time(s) $\downarrow$\\
\midrule
FLSW-100 & \textbf{0.000} & 0.177& 0.903 & 0.433 & 1.76 & 0.013 & 0.147 & 0.914 & 0.434 & 1.31\\
FLSW-200  & \textbf{0.000} & 0.110& 0.941 & 0.475 & \textbf{1.75} & 0.007 & 0.093 & 0.948 & 0.461 & 1.29\\
FLSW-300  & 0.003 & 0.130& 0.929 & 0.399 & 1.76 & 0.007& 0.120 & 0.933 & 0.369 & 1.29 \\
FLSW-400  & 0.003 & 0.153& 0.916 & 0.314 & 1.76 & \textbf{0.003} & 0.150 & 0.917 & 0.288 & 1.27\\
\midrule
\cellcolor{gray!25}WaterSeeker  & \cellcolor{gray!25}0.010& \cellcolor{gray!25}\textbf{0.057}& \cellcolor{gray!25}\textbf{0.966}& \cellcolor{gray!25}\textbf{0.649}& \cellcolor{gray!25}1.89 &\cellcolor{gray!25}0.010& \cellcolor{gray!25}\textbf{0.057}& \cellcolor{gray!25}\textbf{0.966}& \cellcolor{gray!25}\textbf{0.542} & \cellcolor{gray!25}\textbf{0.83}\\

\bottomrule
\end{tabular}
}
\label{tab:more_seg}
\end{table*}

\noindent\textbf{WinMax's Limitations for Multiple Segments}. As shown in Equation \ref{eq:winmax}, when multiple watermarked segments are inserted within the same document, WinMax cannot function properly. In contrast, WaterSeeker is able to adapt to this situation. The detection and localization performance for multiple watermarked segments can be found in the Table \ref{tab:more_seg}, comparing performance of WaterSeeker and FLSW.

\subsection{WaterSeeker Compared with FLSW}
\textbf{Main Experiment}. As shown in Table \ref{tab:main}, although the time cost of FLSW is comparable to that of WaterSeeker, its detection performance is significantly inferior. This difference is due to FLSW's fixed-length nature, which restricts its capability to utilize the gold index for detecting watermarked segments of varying lengths. Results for Mistral-7b is shown in Appendix \ref{sec:appendix_main_mistral}.

\vspace{3pt}

\noindent\textbf{Multiple Segments Detection}. Table \ref{tab:more_seg} presents the detection and localization results for documents containing three watermarked segments. The experimental setup mirrors that of the main experiment, utilizing Llama-2-7b as the generation model. The results indicate that as the number of inserted segments increases, it becomes easier to detect a watermarked segment (all methods show improved F1 scores). However, the Intersection over Union (IoU) did not exhibit significant changes. Notably, WaterSeeker continues to outperform the FLSW algorithm across all four configurations, consistent with the main experiment that included only one segment per sample.

\vspace{3pt}

\noindent\textbf{Further Analysis of Fix-length Nature}. To better illustrate FLSW's inability to adapt to watermarked segments of varying lengths, we selected two types of extreme examples from the main experiment dataset. As shown in Table \ref{tab:flsw}, for segments with strong watermark intensity but short length, using larger window sizes such as FLSW-300 or FLSW-400 leads to the inclusion of many non-watermarked segments, which dilutes the watermark intensity and results in a significant drop in performance. Conversely, for segments with weak watermark intensity but longer length, using smaller window sizes like FLSW-100 or FLSW-200 results in an insufficient number of watermark tokens for detection, preventing the accumulation of intensity and adversely affecting the detection results.

\begin{table}[t]
\centering
\caption{Performance of WaterSeeker and FLSW in two types of examples: segments with strong watermark intensity but short length (length < 150, KGW $\delta=2.0$), and segments with weak watermark intensity but long length (length > 350, KGW $\delta=1.0$).}
\centering
\resizebox{0.3\textwidth}{!}{
\begin{tabular}{lcc}
\toprule
& \textbf{TPR} & \textbf{IoU}   \\ 
\midrule
\multicolumn{3}{c}{\cellcolor{gray!25}\textbf{\textit{Strong but Short}}}\\
FLSW-300 & 0.000 & 0.000 \\ 
FLSW-400 & 0.000 & 0.000 \\ 
WaterSeeker & \textbf{0.667} & \textbf{0.572} \\ 
\midrule
\multicolumn{3}{c}{\cellcolor{gray!25}\textbf{\textit{Weak but Long}}}\\
FLSW-100 & 0.375 & 0.150\\ 
FLSW-200 & 0.625 & 0.447 \\ 
WaterSeeker & \textbf{0.813} & \textbf{0.642} \\ 
\bottomrule
\end{tabular}
}
\label{tab:flsw}
\end{table}

\subsection{Ablation Study}
\label{sec:ablation}
\begin{table}[t]
\caption{This table shows the contributions of the first stage of WaterSeeker: Suspicious Segment Localization. It lists the average coverage of localization results compared to ground truth segments for various watermark algorithms and strengths, along with the average offsets of the detected indices.}
\centering
{
\begin{tabular}{cccc}
\toprule
  & \textbf{Strength} & \textbf{Avg. Cov.} & \textbf{Avg. Off.}\\
 \midrule
 \multirow{3}{*}{\textbf{KGW}} & $\delta=2.0$ & 0.989 & 0.34$W$\\
 & $\delta=1.5$ & 0.964 & 0.35$W$ \\
 & $\delta=1.0$ & 0.950 & 0.43$W$ \\
 \midrule
 \multirow{3}{*}{\textbf{Aar}} & $\text{temp}=0.5$ & 0.945 & 0.13$W$ \\
 & $\text{temp}=0.4$ & 0.948 & 0.12$W$ \\
 & $\text{temp}=0.3$ & 0.920 & 0.10$W$\\
 \bottomrule
\end{tabular}
}
\label{tab:coverage}
\end{table}
\begin{table}[t]
    \caption{Comparison of detection performance with and without Local Traverse Detection.}
    \centering
    \begin{tabular}{ccccc}
        \toprule
        \multirow{2}{*}{\textbf{Settings}} & \multicolumn{2}{c}{\textbf{KGW}} & \multicolumn{2}{c}{\textbf{Aar}} \\
        \cmidrule{2-5}
        & F1 & IoU & F1 & IoU \\
        \midrule
        w. Traversal & \textbf{0.872} & \textbf{0.675} & \textbf{0.819} & \textbf{0.578} \\
        w/o Traversal & 0.817 & 0.576 & 0.765 & 0.509 \\
        \bottomrule
    \end{tabular}
    \label{tab:ablation}
\end{table}
We analyze the effectiveness of the two stages of WaterSeeker through an ablation study. 

\vspace{3pt}

\noindent\textbf{Stage 1: Suspicious Region Localization}. This stage aims to narrow down the detection target from a large document to a smaller region. The goal is to achieve high coverage of the ground truth segments while maintaining the start and end offsets within a specified window size. This ensures that subsequent local traversals can access the gold index. As shown in Table \ref{tab:coverage}, Step 1 achieves an average coverage exceeding 0.9, with average start and end offsets remaining below $W$ across various watermark algorithms and strengths, demonstrating good adaptability.

\vspace{3pt}

\noindent\textbf{Stage 2: Local Traverse Detection}. Local Traverse Detection performs a localized iteration based on the segments from Stage 1, allowing for more refined verification within the window. Table \ref{tab:ablation} shows that across different watermarking algorithms, Local Traverse consistently enhances detection F1 score and average IoU compared to directly applying detection with the localization results, making it an indispensable component of WaterSeeker.

\subsection{Robustness against Text Edit Attacks}
\begin{figure}[t]
    \centering
    \includegraphics[width=\linewidth]{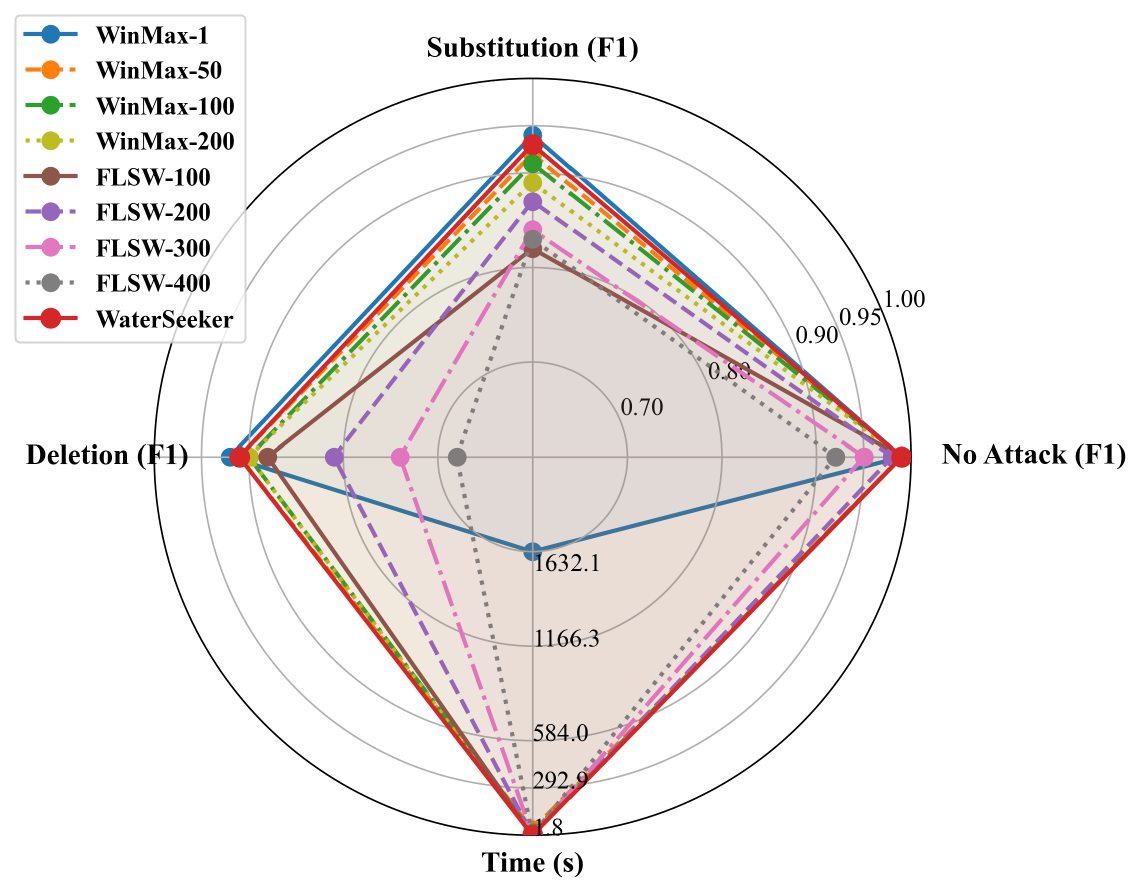}
    \caption{Robustness of WinMax, FLSW, and WaterSeeker against text edit attacks. The detection F1 score is reported for no attack, random word deletion attack (ratio = 0.3), and random word substitution attack (ratio = 0.3), along with the time cost for processing one sample.}
    \label{fig:robustness}
\end{figure}

In this section, we further examine WaterSeeker's robustness against text edit attacks, as watermarked segments generated by LLMs may be altered before integration into non-watermarked documents. 

Figure \ref{fig:robustness} illustrates the detection robustness of WaterSeeker, WinMax and FLSW against random word deletion (ratio=0.3) and substitution attacks (ratio=0.3, utilizing WordNet \citep{fellbaum1998wordnet} for synonym sets). The figure also contrasts these results with those obtained under no attack, along with the associated time costs. In this experiment, the KGW watermarking algorithm was employed, and Llama-2-7b is utilized as generation model. Since text editing can weaken the strength of the watermark, the watermark fragment intensity was set to a strong level under the "No attack" condition, specifically $\delta = 2.0$.

It can be observed from the figure that WaterSeeker achieves strong robustness against word deletion and word substitution attacks, with an F1 score exceeding 0.9. Compared with other baselines, only WinMax-1 perform slightly better than WaterSeeker, but at a significant cost in terms of time.

\section{Conclusion}
This work introduces a new scenario for detecting watermarked segments in large documents and establishes corresponding evaluation metrics. We identified the limitations of full-text detection methods in this context and proposed a ``first locate, then detect" watermark detection algorithm that utilizes a coarse-to-fine strategy. We validated the detection performance and time complexity of our algorithm through a series of analyses and experiments, demonstrating its ability to effectively balance both aspects. Future research could explore more advanced locating methods based on this concept to potentially yield improved detection results.
\section*{Limitations}
While our method has demonstrated effectiveness in detecting watermarked segments within large documents, there are still some limitations that need to be addressed in future work. First, from an evaluation perspective, due to resource constraints, we only conducted experiments on Llama-2-7B and Mistral-7B models. The effectiveness of our method on larger and more powerful models remains to be further verified. Second, WaterSeeker's performance may decrease with very short or weak watermarks. Enhancing the sensitivity of WaterSeeker to detect shorter and weaker watermarks is an area for future improvement, which may involve refining the anomaly extraction algorithms or incorporating additional contextual analysis. Lastly, parameter selection, including threshold settings to control false positives, is crucial and can be challenging in different environments. Stricter threshold controls can reduce the detection rate, necessitating adjustments based on the specific requirements of the actual settings.
\section*{Acknowledgments}
This work is primarily supported by the Key Research and Development Program of China (No. 2024YFB3309702). We would like to express our gratitude to the anonymous ARR October reviewers (Reviewer 6E27, 1X1q, B2fg) and Area Chair zwno for their valuable feedback and suggestions that helped improve this paper. Additionally, we extend our sincere thanks to Hanyu Xue for his assistance with mathematical derivations, which provided valuable insights throughout this research.

\bibliography{custom}

\appendix
\onecolumn
\newpage

\section{Details of Representative Watermarking Algorithms}
\label{sec:appendix_scheme}
\subsection{KGW}
\textbf{Watermarking.} In watermarked text generation, the process for the $t$-th token begins by hashing preceding tokens with a secret key, creating a red-green vocabulary partition where green tokens comprise a fraction $\gamma$. Green token logits are then incrementally increased by $\delta$, which can be expressed as follows:
\begin{equation}
    l'_t(y) = \begin{cases}
        l_t(y), & y \in R_t \\
        l_t(y) + \delta, & y \in G_t
    \end{cases}
\end{equation}
This subtle modification results in watermarked text exhibiting a higher frequency of green tokens compared to non-watermarked text.

\vspace{5pt}

\noindent\textbf{Detection.} Detecting a KGW watermark entails computing red-green partitions for each position using preceding tokens and the hash function, then calculating the green token proportion using the z-score:
\begin{equation}
    z = \frac{|s|_G - \gamma N}{\sqrt{\gamma(1 - \gamma) N}}
\end{equation}
, where \( |s|_G \) represents the total count of green tokens in the whole text of length $N$.

\subsection{Aar}
\textbf{Watermarking.} When generating the $t$-th token, it first involves hashing the preceding tokens using a secret key to obtain a pseudo vector $u_t \sim \text{Uniform}([0,1])^{|V|}$. The $t$-th token is determined by
\begin{equation}
\mathop{\arg\max}\limits_{y}\, u_t(y)^{1/p_t(y)}, 
\end{equation}
where $p$ is the probability vector produced by LLM at the $t$-th step. 
Let's perform equivalent transformations on it:
\begin{align}
\bf{y} &= \mathop{\arg\max}\limits_{y}\, u_t(y)^{1/p_t(y)} \\
&= \mathop{\arg\max}\limits_{y}\, \frac{1}{p_t(y)}\log u_t(y)\\
&= \mathop{\arg\min}\limits_{y}\, \frac{1}{p_t(y)}\log \frac{1}{u_t(y)} \\
&= \mathop{\arg\min}\limits_{y}\, \log\frac{1}{p_t(y)}+ \log\log\frac{1}{u_t(y)} \\
&= \mathop{\arg\max}\limits_{y}\, \log p_t(y) - \log\log\frac{1}{u_t(y)} \label{eq:gumbel}
\end{align}
Given that the probabilistic output $p_t$ of an LLM is derived from the logits $l_t$ through a softmax transformation, and when we additionally consider the sampling temperature $T$, Equation \ref{eq:gumbel} becomes equivalent to:
\begin{equation}
\mathop{\arg\max}\limits_{y}\, \cfrac{l_t(y)}{T} + G_t(y),
\end{equation}
where $l_t$ is the logits produced by the LLM, and $G_t$ is the Gumbel noise: $G_t(y) \sim Gumbel(0,1)$. The $Gumbel(0,1)$ distribution is defined as follows: if $u \sim \text{Uniform}(0,1)$, then $-\log(-\log(u))$ follows a $Gumbel(0,1)$ distribution. 

It is evident that the temperature $T$ can be utilized to exert control over the watermark strength. As the value of $T$ increases, the influence of Gumbel noise on the sampling process becomes more pronounced, consequently resulting in a stronger watermark.

\vspace{5pt}

\noindent\textbf{Detection.} Detecting an Aar watermark involves calculating the correlation value between the pseudo vector $u_t$ and the corresponding token $y_t$ in the text to be examined. The correlation value can be expressed as:
\begin{equation}
    \log \cfrac{1}{1-u_t(y_t)}.
\end{equation}
For the entire text, the statistic value can be expressed as:
{\small
\begin{equation}
\text{p-value} = \Gamma\left(\sum_{t=1}^{N} \log\left(\frac{1}{1 - u_t(y_t)}\right), N, \text{loc}=0, \text{scale}=1\right),
\end{equation}
}
where $\Gamma$ is the Gamma Transformation function that converts the sum of correlation values to a p-value.

\section{Supplementary Experimental Results Using Mistral-7b}
\label{sec:appendix_main_mistral}
\begin{table*}[t]
\caption{We evaluated the detection performance of WaterSeeker against various methods, including Full-text Detection, WinMax \citep{kirchenbauer2023reliability}, and FLSW. Results in this table utilized Mistral-7b as the generation model. The metrics reported include false positive rate (FPR), false negative rate (FNR), F1 score, average Intersection over Union (IoU) between detected and ground truth segments, and processing time per sample. Best performances are highlighted in bold, while the second-best are underlined.}
\centering
\resizebox{0.98\textwidth}{!}{
\begin{tabular}{lccccc|ccccc}
\toprule
\multirow{2}{*}{\textbf{Method}} & \multicolumn{5}{c}{\textbf{KGW}} & \multicolumn{5}{c}{\textbf{Aar}} \\
\cmidrule{2-11}
& FPR$\downarrow$ & FNR$\downarrow$ & F1$\uparrow$ & IoU$\uparrow$ & Time(s)$\downarrow$ & FPR$\downarrow$ & FNR$\downarrow$ & F1$\uparrow$ & IoU$\uparrow$ & Time(s)$\downarrow$ \\
\midrule
WinMax-1 & 0.010 & \textbf{0.243} &  \textbf{0.857} & \textbf{0.641} & 1632.11 & 0.013 &\textbf{0.297} & \textbf{0.819} & \textbf{0.588} & 3615.42\\
WinMax-50  & 0.007 & 0.270 & 0.841 & 0.613 & 34.31 & 0.007 & 0.320 & 0.806 & 0.560& 72.13\\
WinMax-100  & \underline{0.003} & 0.283 & 0.833 & 0.588 & 17.16 & 0.007 & 0.333 & 0.797 & 0.531 & 35.34\\
WinMax-200  & \underline{0.003} & 0.340 & 0.794 & 0.511 & 9.12 & \underline{0.003} & 0.347 & 0.789 & 0.501 & 18.38\\
\midrule
\cellcolor{gray!25}WaterSeeker  & \cellcolor{gray!25}0.010& \cellcolor{gray!25}\underline{0.253}& \cellcolor{gray!25}\underline{0.850}& \cellcolor{gray!25}\underline{0.634}& \cellcolor{gray!25}\textbf{1.75} & \cellcolor{gray!25}0.010& \cellcolor{gray!25}\underline{0.300}& \cellcolor{gray!25}\textbf{0.819}& \cellcolor{gray!25}\underline{0.563}& \cellcolor{gray!25}\textbf{0.41}\\
\midrule
FLSW-100 & \textbf{0.000} & 0.463 & 0.698 & 0.393 & \underline{1.76} & \textbf{0.000} & 0.473 & 0.690 & 0.367 & 1.31\\
FLSW-200  & \textbf{0.000} & 0.373 & 0.770 & 0.426 & \underline{1.76} & \underline{0.003} & 0.387 & 0.759 & 0.412 & 1.29 \\
FLSW-300 & \underline{0.003} & 0.437 & 0.719 & 0.331 & \underline{1.76} & \underline{0.003} & 0.387 & 0.759& 0.334& 1.29 \\
FLSW-400  & \underline{0.003} & 0.540 & 0.629 & 0.218 & \textbf{1.75} & 0.007 & 0.440 & 0.715& 0.249& \underline{1.27}\\

\bottomrule
\end{tabular}
}
\label{tab:main_appendix}
\end{table*}
Supplementary results for main experiment using Mistral-7b as generation model is shown in Table \ref{tab:main_appendix}, showing consistent trend with Llama-2-7b.

\section{Pseudocode of Detection Baselines}

\label{sec:appendix_baselines}
Pseudocode of WinMax and FLSW could be found in Algorithm \ref{alg:winmax} and \ref{alg:flsw}, respectively.

\begin{algorithm}
\caption{WinMax Algorithm}
\begin{algorithmic}[1]
\Procedure{WinMaxDetection}{tokens, interval, threshold}
    \State hasWatermark $\gets$ False, indices $\gets$ [ ]
    \State maxStat $\gets$ - $\infty$, bestIndex $\gets$ None
    \For{$W$ $\in$ [1, len(tokens)], step=interval}
        \For{$i$ \textbf{ in } 0 \textbf{ to } $\text{len(tokens)} - W$}
            \State stat $\gets$ WatermarkScore(tokens[$i:i+W$])
            \If{stat $>$ maxStat}
                \State maxStat $\gets$ stat
                \State bestIndex $\gets$ $(i, i + W)$
            \EndIf
        \EndFor
    \EndFor
    \If{maxStat $>$ threshold}
        \State hasWatermark $\gets$ True
        \State indices.append(bestIndex)
    \EndIf
    \State\Return hasWatermark, indices
\EndProcedure
\end{algorithmic}
\label{alg:winmax}
\end{algorithm}

\begin{algorithm}
\caption{FLSW Algorithm}
\begin{algorithmic}[1]
\Procedure{FLSWDetection}{tokens, $W$, threshold}
    \State hasWatermark $\gets$ False
    \State indices $\gets$ [ ]
    \For{$i$ \textbf{ in } 0 \textbf{ to } $\text{len(tokens)} - W$}
        \State stat $\gets$ WatermarkScore(tokens[$i:i+W$])
        \If{stat $>$ threshold}
            \State hasWatermark $\gets$ True
            \State indices.append(($i$, $i + W$))
        \EndIf
    \EndFor
    \State indices $\gets$ ConnectFragments(indices)
    \State\Return hasWatermark, indices
\EndProcedure
\end{algorithmic}
\label{alg:flsw}
\end{algorithm}

\section{Detail of Threshold Selection Within the Specified Window}
\label{sec:appendix_threshold}
A key role of threshold selection is to control the false positive rate. In this context, the task involves detecting watermark fragments within long texts, which requires traversing extensive content and can lead to an accumulation of false positives. Therefore, managing the false positive rate within the detection window is crucial in this scenario. In the experiment, we set the target false positive rate $\alpha$ within the detection window to $10^{-6}$.

\subsection{Rationale for setting $\alpha$ to $10^{-6}$}

\begin{table}[h!]
  \centering
  \caption{Simulated FPR of WaterSeeker using 10,000 samples for each watermarking method. The targeted false positive rate within the detection window is set to $10^{-6}$.}
  \begin{tabular}{cc}
    \toprule
    Watermarking Method & Simulated FPR \\
    \midrule
    KGW & 0.0054 \\
    Aar & 0.0042 \\
    \bottomrule
  \end{tabular}
  \label{tab:simulated_fpr}
\end{table}

WaterSeeker, WinMax, and FLSW all involve employing sliding windows for text traversal and conduct full-text detection within each window. As these windows overlap, they cannot be treated as independent, making it challenging to derive a theoretical upper bound for the document-level FPR from the target FPR within each window. Given this, we utilize large-scale data simulation to demonstrate that, with a target false positive rate of $10^{-6}$ within each window, our proposed method WaterSeeker maintains an acceptable false positive rate.

For the KGW method, we set $\gamma=0.5$ in our experiments, meaning each token in non-watermarked text has a 0.5 probability of being green and 0.5 probability of being red. In the simulation, we generate 10,000 samples, each containing 10,000 tokens, with each token having a 0.5 probability of being 1 and 0.5 probability of being 0. For the Aar method, each token in non-watermarked text corresponds to $u_i \sim \text{Uniform}[0,1]$. In the simulation, we again generate 10,000 samples, each containing 10,000 tokens, with each token randomly assigned a floating-point number from $[0,1]$.

We then apply WaterSeeker to detect watermarked segments within these samples, setting the target false positive rate within the detection window to $10^{-6}$. The large-scale simulation results in Table \ref{tab:simulated_fpr} demonstrate that WaterSeeker maintains a false positive rate of approximately 0.005, which is considered acceptable. For scenarios requiring more stringent FPR control, the target false positive rate can be adjusted downward. However, this inevitably compromises the detection rate, highlighting a key challenge in watermarked segment detection within large documents.

\subsection{Setting the threshold to achieve a target false positive rate $\alpha$}

\textbf{KGW}. For KGW, as analyzed in Section \ref{sec:theory}, when the window size is large, we can approximate using the Central Limit Theorem, resulting in \( z^* = \Phi^{-1}(1 - \alpha) \). When \( \alpha = 10^{-6} \), this gives \( z \approx 4.75 \). However, when the window size \( W \) is small, the approximation to a normal distribution using the Central Limit Theorem may lead to significant deviations. Therefore, we will use the binomial distribution for precise calculations. $x \sim B(W, \gamma)$ describes the number of green tokens in a window of size $W$ follows a binomial distribution, therefore:

\[
z = \frac{x - \gamma W}{\sqrt{W \gamma (1 - \gamma)}}.
\]

To find \( P(z \geq z^*) \):

\[
P(z \geq z^*) = P\left(\frac{x - \gamma W}{\sqrt{W \gamma (1 - \gamma)}} \geq z^*\right).
\]

Expanding this, we have:

\[
P(z \geq z^*) = \sum_{k=0}^{W} \binom{W}{k} \gamma^k (1 - \gamma)^{W - k} \mathbb{I}\left\{\frac{k - \gamma W}{\sqrt{W \gamma (1 - \gamma)}} \geq z^*\right\}.
\]

This is the exact expression for \( P(z \geq z^*) \) without any approximations.

We can further simplify:

$$
P(z \geq z^*) = \sum_{k=0}^{W} \binom{W}{k} \gamma^k (1 - \gamma)^{W - k} \mathbb{I}\left\{k \geq \gamma W + z^* \sqrt{W (1 - \gamma)}\right\}.
$$

We need to find an appropriate \( z^* \) such that \( P(z \geq z^*) < \alpha \). This function does not have a direct analytical solution, so we can increment \( z^* \) in steps of 0.01 until the probability exceeds \( \alpha \). The final value of \( z^* \) is dependent on \( W \), and we pre-compute these values during experiments and store them in a dictionary. In experiments, for detected segments with a length of 200 or more, we directly apply the Central Limit Theorem approximation, setting $z = 4.75$. For segments shorter than 200, we use the binomial distribution and retrieve the corresponding threshold from the pre-computed dictionary.

\vspace{5pt}

\noindent\textbf{Aar}. For Aar, recall the p-value calculation formula:
\begin{equation}
\text{p-value} = \Gamma(S, W, \text{loc}=0, \text{scale}=1),
\label{eq:aar}
\end{equation}

where $S = \sum_{i=1}^{W} \log(\frac{1}{1-u_i})$, and $W$ is the window size. For non-watermarked text, $u_i \sim \text{Uniform}([0,1])$. Consequently, $S$ follows a Gamma distribution: $S \sim \text{Gamma}(W, 1)$, where $W$ is the shape parameter and 1 is the scale parameter. Equation \ref{eq:aar} is equivalent to:
\begin{equation}
    \text{p-value} = 1 - \text{GammaCDF}(S, W, 1),
\end{equation}
where GammaCDF is the cumulative distribution function of the Gamma distribution with shape parameter $W$ and scale parameter 1. To achieve a false positive rate of $\alpha$, we need to set a threshold $p^*$ such that: $P(\text{p-value} < p^*) = \alpha$. Given the definition of p-value, this is equivalent to:
$P(1 - \text{GammaCDF}(S, W, 1) < p^*) = \alpha$, which can be rewritten as: $P(S > \text{GammaInv}(1 - p^*, W, 1)) = \alpha$, where GammaInv is the inverse of the Gamma CDF. Since $S$ follows a $\text{Gamma}(W, 1)$ distribution for non-watermarked text, we can express this as:
\begin{equation}
   1 - \text{GammaCDF}(\text{GammaInv}(1 - p^*, W, 1), W, 1) = \alpha.
\end{equation}

Solving this equation for $p^*$, we get $p^* = \alpha$, which is also a constant value for different $W$.

\section{Impact of Window Size on Watermark Intensity Calculation}
\label{sec:appendix_window_size}

\begin{figure}[t]
    \centering
    \begin{subfigure}[b]{\textwidth}
        \centering
        \includegraphics[width=0.95\textwidth]{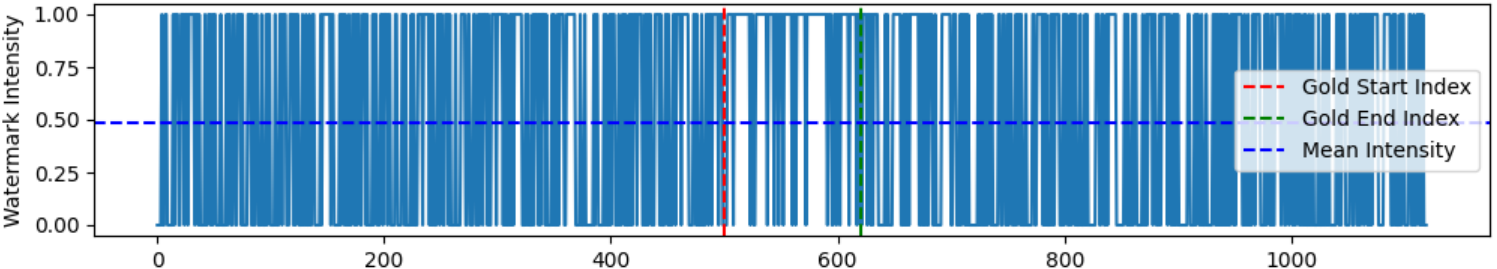}
        \caption{$W=1$}
        \label{fig:kgw_intensity_w_1}
    \end{subfigure}
    
    \vspace{1em}
    
    \begin{subfigure}[b]{\textwidth}
        \centering
        \includegraphics[width=0.95\textwidth]{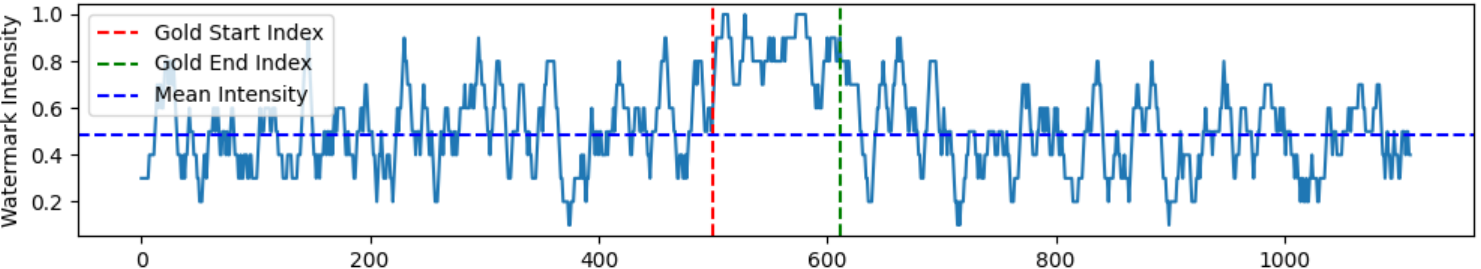}
        \caption{$W=10$}
        \label{fig:kgw_intensity_w_10}
    \end{subfigure}
    
    \vspace{1em}
    
    \begin{subfigure}[b]{\textwidth}
        \centering
        \includegraphics[width=0.95\textwidth]{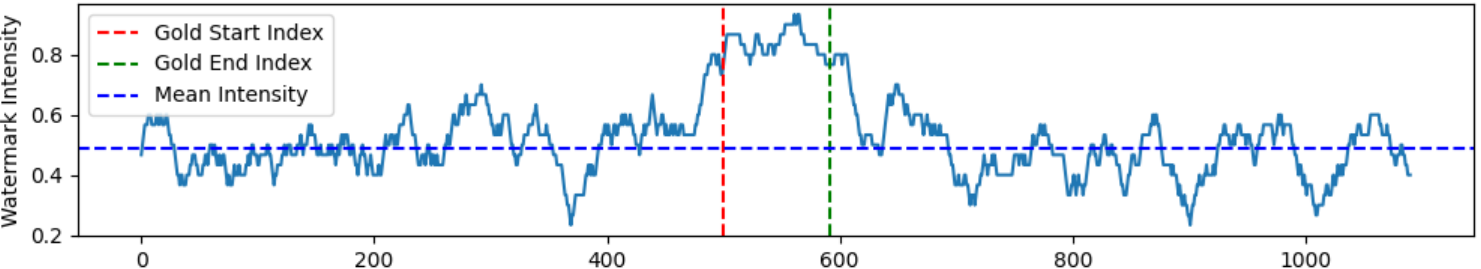}
        \caption{$W=30$}
        \label{fig:kgw_intensity_w_30}
    \end{subfigure}
    
    \vspace{1em}
    
    \begin{subfigure}[b]{\textwidth}
        \centering
        \includegraphics[width=0.95\textwidth]{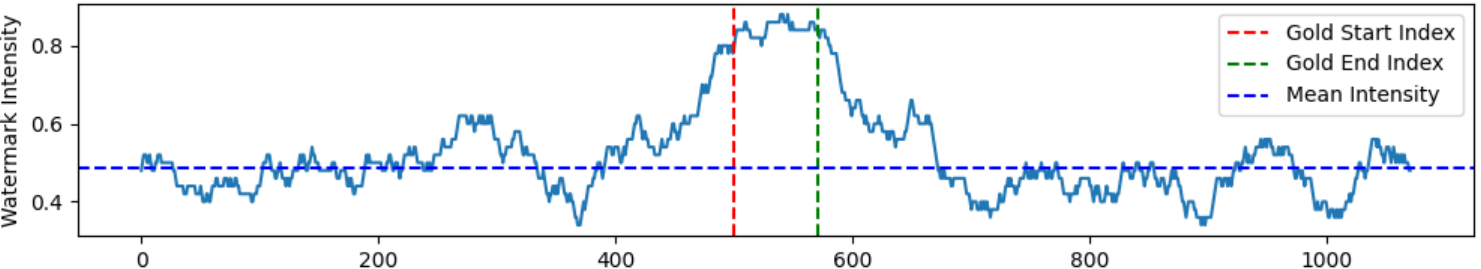}
        \caption{$W=50$}
        \label{fig:kgw_intensity_w_50}
    \end{subfigure}
    
    \caption{Case study: Impact of varying window sizes on watermark intensity calculation in the KGW algorithm.}
    \label{fig:kgw_intensity}
\end{figure}

\begin{figure}[t]
    \centering
    \begin{subfigure}[b]{\textwidth}
        \centering
        \includegraphics[width=0.95\textwidth]{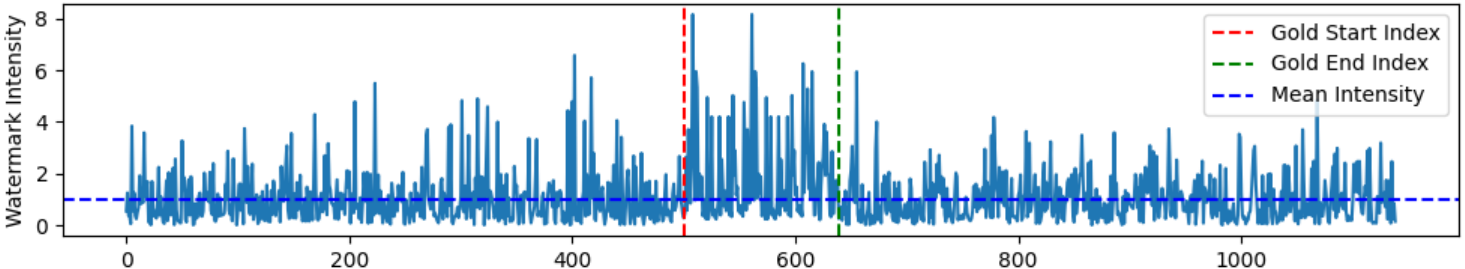}
        \caption{$W=1$}
        \label{fig:aar_intensity_w_1}
    \end{subfigure}
    
    \vspace{1em}
    
    \begin{subfigure}[b]{\textwidth}
        \centering
        \includegraphics[width=0.95\textwidth]{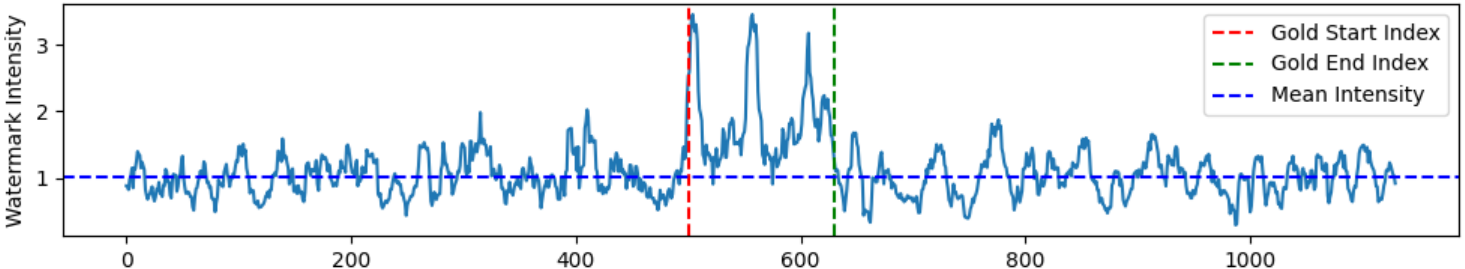}
        \caption{$W=10$}
        \label{fig:aar_intensity_w_10}
    \end{subfigure}
    
    \vspace{1em}
    
    \begin{subfigure}[b]{\textwidth}
        \centering
        \includegraphics[width=0.95\textwidth]{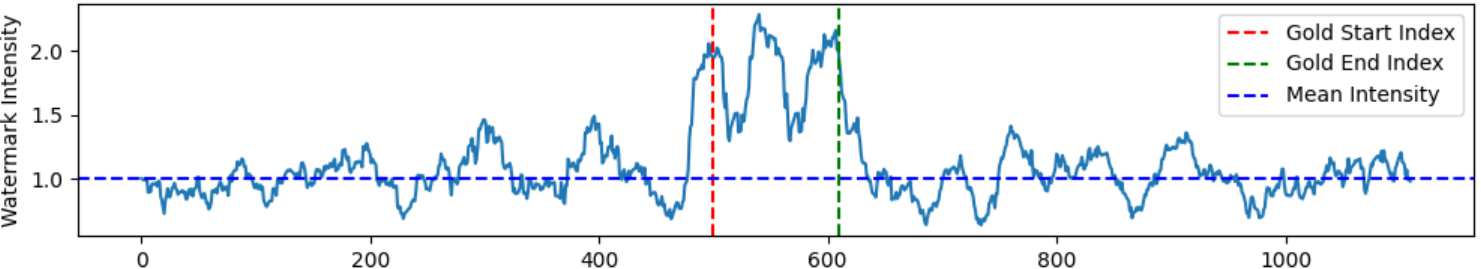}
        \caption{$W=30$}
        \label{fig:aar_intensity_w_30}
    \end{subfigure}
    
    \vspace{1em}
    
    \begin{subfigure}[b]{\textwidth}
        \centering
        \includegraphics[width=0.95\textwidth]{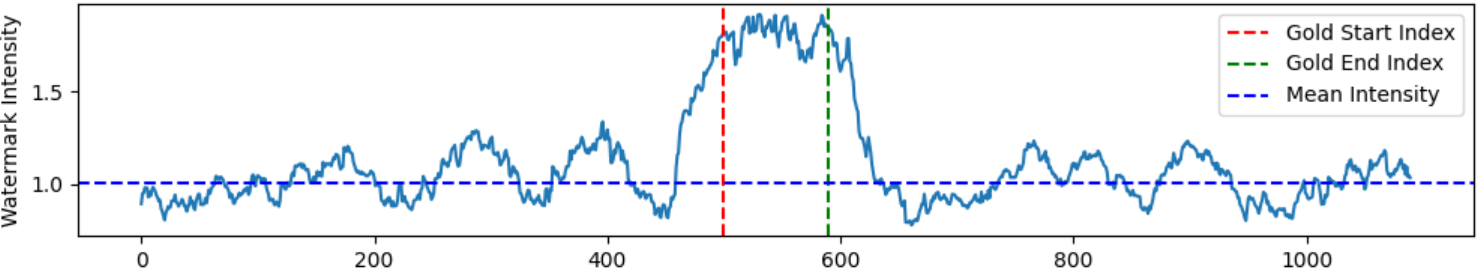}
        \caption{$W=50$}
        \label{fig:aar_intensity_w_50}
    \end{subfigure}
    
    \caption{Case study: Impact of varying window sizes on watermark intensity calculation in the Aar algorithm.}
    \label{fig:aar_intensity}
\end{figure}

The first step in WaterSeeker is score list computation. In this step, selecting an appropriate window size $W$ for calculating mean scores is crucial. A small $W$ introduces excessive noise, while a large $W$ reduces granularity and increases computational time due to the need to examine $W^2$ windows during local traversal. Therefore, we aim to determine an appropriate window size that is relatively small while still providing a sufficiently smooth representation of watermark intensity throughout the text.

We present a case study comparing watermark intensity calculations using window sizes $W=1, 10, 30,$ and $50$. The analysis encompasses the ground truth segment and 500 tokens on either side. Figures \ref{fig:kgw_intensity} and \ref{fig:aar_intensity} illustrate the results for the KGW and Aar algorithms, respectively. The intensity curves reveal that small window sizes, particularly $W \leq 10$, introduce significant fluctuations. While $W=30$ exhibits reduced noise, it still presents instabilities, as shown in Figure \ref{fig:aar_intensity_w_30} (the ground truth segment part). Overall, $W=50$ demonstrates the least noise. Consequently, we adopt $W=50$ for our main experiments.

\end{document}